\documentclass[10pt]{article} 
\makeatletter
\def\blfootnote{\gdef\@thefnmark{}\@footnotetext}
\makeatother

\usepackage[preprint]{tmlr}

\usepackage{amsmath,amsfonts,bm}

\def\eqref#1{equation~\ref{#1}}

\def\1{\bm{1}}

\DeclareMathAlphabet{\mathsfit}{\encodingdefault}{\sfdefault}{m}{sl}
\SetMathAlphabet{\mathsfit}{bold}{\encodingdefault}{\sfdefault}{bx}{n}

\usepackage{hyperref}
\usepackage{url}
\usepackage{graphicx}
\usepackage{amsmath}
\usepackage{amssymb}
\usepackage{booktabs}
\usepackage{color}
\usepackage{colortbl}
\usepackage{multirow}
\usepackage{tabu}
\usepackage{xcolor}
\usepackage{pifont}
\usepackage{subcaption}
\usepackage{layouts}
\usepackage{makecell}
\usepackage{soul}

\usepackage{tabularx}
\usepackage{amsmath,amssymb}

\usepackage{dsfont}
\usepackage{amsmath,amssymb}

\newcommand\extrafootertext[1]{
    \bgroup
    \renewcommand\thefootnote{\fnsymbol{footnote}}
    \renewcommand\thempfootnote{\fnsymbol{mpfootnote}}
    \footnotetext[0]{#1}
    \egroup
}

\definecolor{LightCyan}{rgb}{0.318, 0.439, 0.843}
\definecolor{LightCyan}{rgb}{0.8295, 0.85975, 0.96075}
\newcolumntype{g}{>{\columncolor{LightCyan}}c}
\definecolor{IncrGreen}{rgb}{0.25, 0.55, 0.25}
\definecolor{DecrRed}{rgb}{0.55, 0.25, 0.25}

\title{
  DreamPaint: Few-Shot Inpainting of E-Commerce Items \\ for Virtual Try-On without 3D Modeling
}

\author{\name Mehmet Saygin Seyfioglu $^\dag$ \email msaygin@cs.washington.edu \\
      \addr \emph{University of Washington}
      \AND
      \name Karim Bouyarmane \email bouykari@amazon.com \\
      \addr \emph{Amazon}
      \AND
      \name Suren Kumar \email ssurkum@amazon.com\\
      \addr \emph{Amazon}
      \AND
      \name Amir Tavanaei \email atavanae@amazon.com\\
      \addr \emph{Amazon}
      \AND
      \name Ismail B. Tutar \email ismailt@amazon.com\\
      \addr \emph{Amazon}}

\def\month{MM}  
\def\year{YYYY} 
\def\openreview{\url{https://openreview.net/forum?id=XXXX}} 
\begin{document}

\maketitle
\blfootnote{$^\dag$ Work done during an internship at Amazon}

\begin{abstract}
We introduce DreamPaint, a framework to intelligently inpaint any e-commerce product on any user-provided context image. The context image can be, for example, the user’s own image for virtual try-on of clothes from the e-commerce catalog on themselves, the user’s room image for virtual try-on of a piece of furniture from the e-commerce catalog in their room, etc. As opposed to previous augmented-reality (AR)-based virtual try-on methods, DreamPaint does not use, nor does it require, 3D modeling of neither the e-commerce product nor the user context. Instead, it directly uses 2D images of the product as available in product catalog database, and a 2D picture of the context, for example taken from the user's phone camera. The method relies on few-shot fine tuning a pre-trained diffusion model with the masked latents (e.g., Masked DreamBooth) of the catalog images per item, whose weights are then loaded on a pre-trained inpainting module that is capable of preserving the characteristics of the context image. DreamPaint allows to preserve both the product image and the context (environment/user) image without requiring text guidance to describe the missing part (product/context). DreamPaint also allows to intelligently infer the best 3D angle of the product to place at the desired location on the user context, even if that angle was previously unseen in the product’s reference 2D images. We compare our results against both text-guided and image-guided inpainting modules and show that DreamPaint yields superior performance in both subjective human study and quantitative metrics.  
\end{abstract}


\section{Introduction} 

A long-standing problem in E-Commerce is the ability of customers to try-on the products before the purchase. The lack of try-on possibility increases the risk and cost associated with product returns due to misfit of the product after the product is delivered and physically tried on. This problem arises across different product categories. For example, for clothes, shoes, jewelry, watches, glasses, etc, the customer might want to try the product on themselves before the purchase. For pieces of furniture like couches, tables, chairs, decorative object, etc, the customers might want to try them on their own room, and own setting, to visualize the product in-context before the purchase. The same applies for multiple other categories of products.

One solution to tackle this problem is to use augmented-reality through 3D modeling. Such solutions, however, typically require a 3D model of the product or an expensive method to reconstruct a 3D model from high quality 2D images. The vast majority of shopping websites catalogs of products do not have 3D models associated with them (either native or reconstructed). As a result, the AR-based virtual try-on capability is only offered 1) on a portion of shopping websites and 2) on a portion of products within general online stores that are not specialized on a single category of products.

\begin{figure*}[t]
    \centering
    \includegraphics[width=\textwidth]{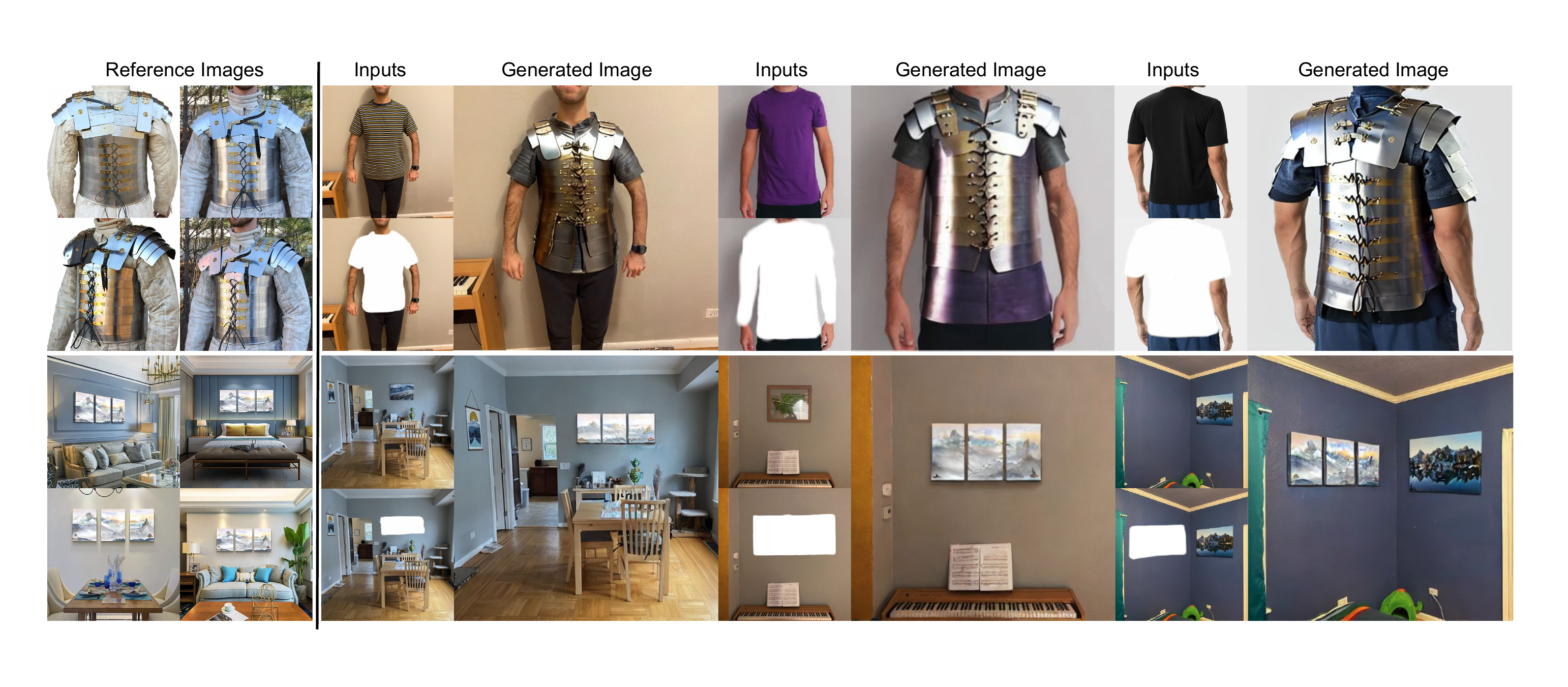}
    \caption{ Some example outputs of our DreamPaint model. We fine-tune the U-Net and the text encoder of the Stable Diffusion model in a few shot setting with the given masked reference images, which allows users to inpaint their personal images to virtually "try" the e-commerce items before buying.}
    \label{fig:examples_intro}
\end{figure*}

We propose a solution for virtual try-on that does not require 3D modeling or augmented reality. It only uses whatever set of 2D images is available for the product in the product catalog database of the shopping website. The method relies on a combination of Masked DreamBooth and Stable Diffusion Inpainting module. We name the method DreamPaint. The user provides a 2D image of themselves (for clothes), their room (for furniture), their desk setting (for decorative objects) etc. The user can then specify on that image the rough location in which they want the product to be placed. DreamPaint positions the product in that specified location seamlessly integrating with the environment. The model is not constrained by the images of the product that are available. From these images, it can intelligently extrapolate new angles if such angles are required by the environment configuration to place the object. Some examples generated by DreamPain can be seen on Fig. \ref{fig:examples_intro}.

Image generation using diffusion-based models has attracted a lot of attention recently. Other diffusion-based approaches to the problem typically preserve only one input: either the object is preserved and the environment surrounding the object is generated with text-guidance (Dreambooth-like) or the environment is preserved and the object within the environment is generated with text-guidance (Inpainting-like). Our approaches combines both in a novel way to preserve both the environment and the object inputs.


The rest of the paper is organized as follows. In Section 2, we review closely related work. In Section 3, we present the DreamPaint method, combining Masked Stable Diffusion Dreambooth and Inpainting module. In Section 4 we show experimental settings. In Section 5 we compare the results against text guided and image guided inpainting models and show that DreamPaint outperforms them in both qualitative and quantitative comparisons as well as a human study. We do some ablations on Section 6 and end our paper with Conclusions and Limitations of our model.

\begin{figure*}[t]
    \centering
    \includegraphics[width=\textwidth]{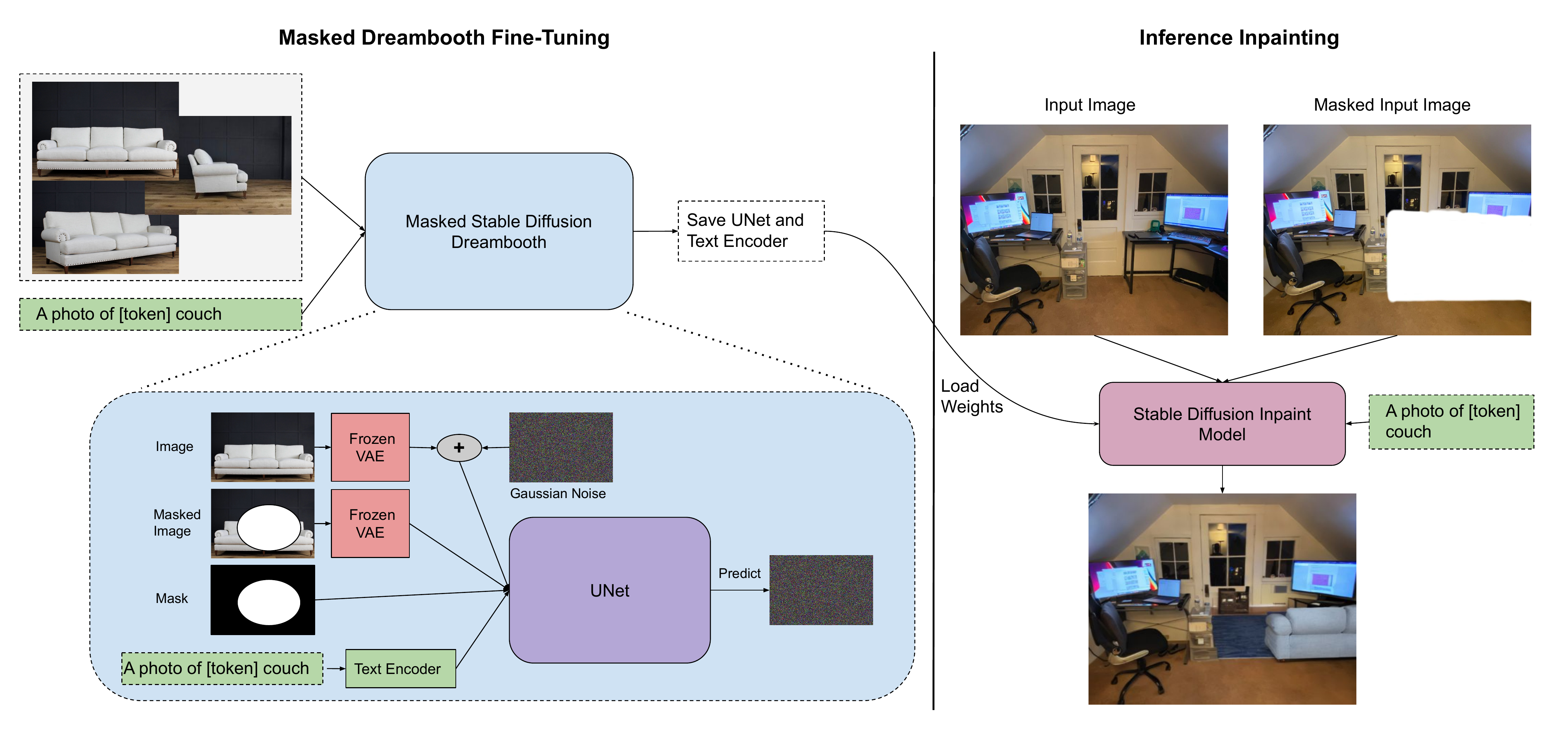}
    \caption{Model overview. Similar to Dreambooth, our model takes a small number of reference images of a specific item taken from different angles and an associated class name (a couch in this example). Then we train the text encoder and U-Net of the Stable Diffusion using randomly masked versions of reference images for the inpainting setting. After the Dreambooth training is done, we load the U-Net and text encoder weights on the Stable Diffusion Inpainting Model, which is capable of inpainting the masked region for the given image. Note that, on the inference setting, the inpainting model generated the item in a novel view. }
    \label{fig:method}
\end{figure*}
\section{Related Work}
Text-to-image diffusion models have shown unprecedented success in recent research \cite{ramesh2022hierarchical}, \cite{saharia2022photorealistic}, \cite{yu2022scaling}, \cite{rombach2022high}. When trained with large-scale datasets of image and text pairs, these models can generate highly accurate, and semantically meaningful images utilizing text prompts, especially for common objects. 


 Diffusion models can also be further trained for personalization tasks such as inpainting \cite{suvorov2022resolution}. As shown in \cite{avrahami2022blended}, natural language provides a good guidance for inpainting of common objects. However, for uncommon objects, such as the items found in an e-commerce catalog, these models generally fail to generate a satisfactory representation of the item by preserving its characteristic details. Moreover, even in the cases where the model has a generative capacity for the given object, text descriptions are highly ambiguous by their nature and are inefficient in conveying the characteristic details of an object. Thus, an image-based diffusion guidance would better serve for inpainting of e-commerce items. 




Paint by Example (PBE) \cite{yang2022paint}, is a recent image-guided diffusion model, which utilizes an exemplar image to guide the diffusion inpainting process. The method achieves superior performance against text-guided inpainting models like Stable Diffusion \cite{rombach2022high} or harmonization models such as DCCF \cite{xue2022dccf} for in-the-wild object inpainting. However, the PBE method also has some drawbacks in preserving the high-fidelity details of objects, especially for the  underrepresented objects, as they embed the exemplar image using only CLIP's CLS embedding for guidance. Relying on such high-level embeddings results in omitting fine-grained details that define the characteristics of many e-commerce items. Therefore, an alternative approach is needed for the virtual try-on setting. 

Another recent technique called DreamBooth~\cite{ruiz2022dreambooth} offers high-fidelity concept learning on novel images. Given a few reference images (best if provided from different angles), a new token representing these reference images could be injected in the model by few-shot fine-tuning the model's denoiser. E-commerce data is a good fit for this approach because every product in e-commerce catalogs typically have multiple images taken from different angles.

Inspired by DreamBooth, we propose \textbf{Dreampaint} (\textbf{Dream}booth-in\textbf{paint}), where we combine DreamBooth and inpainting modules to learn e-commerce objects by preserving their fidelity. By utilizing a pre-trained Stable Diffusion model, to make the new concept learned by Dreambooth suitable for inpainting, we are modifying the Dreambooth approach into masked training few-shot fine-tuning to learn a new U-Net \cite{ronneberger2015u} and text-encoder that are injected with the new item. We then load our fine-tuned U-Net and text-encoder into a pre-trained Stable diffusion inpainting model, which allows the user to mask a portion of their personal images and generate the injected product in the masked region by preserving both the product specific high-fidelity features as well as the context of the user-provided scene. 


Furthermore, DreamPaint can fill in the masked region with a standardized prompt as in \cite{ruiz2022dreambooth}. However, if the results turn out unsatisfactory for the user, the model has the flexibility to be further tuned by additional text prompts. Therefore, in a way, DreamPaint is leveraging both image and text guidance.

We compare our results against 1) the image-guided model PBE (since it was domenstrated to outperform other methods in preserving the fidelity, which is our main objective), and 2) a standard text-guided Stable Diffusion inpainting model that inpaints the given image using the catalog title of the reference item. 
\section{Methodology}
This section is organized in three parts. In the first part we discuss latent diffusion models (\ref{subsection:latent}) and our first baseline - the text-guided inpainting (\ref{subsection:inpainting}). In the second part (\ref{subsection:pbe}), we discuss the example-based painting method. Finally, in the third part, we introduce the Dreambooth method (\ref{subsection:dreambooth}) and we explain our DreamPaint proposed approach for the high-fidelity e-commerce inpainting task (\ref{subsection:dreampaint}). 

\subsection{Latent Diffusion Models}\label{subsection:latent}
Diffusion models are generative models that learn the data distribution by reversing a fixed-length Markovian forward process, thereby iteratively denoising a normally distributed variable \cite{sohl2015deep}. Lately, it is shown that, instead of using the pixel space, denoising can be conducted in a latent space, which is computationally efficient as it reduces the dimension of images, as well as it omits the high frequency noise within the given image. One example of a latent diffusion models is Stable Diffusion \cite{rombach2022high}, which consists of three main components: A Variational Autoencoder (VAE) to transform the given input in a latent space, a text encoder to process the given text, and a time-conditioned U-Net \cite{ronneberger2015u} to predict the noise that is added on the image latents which are conditioned by the text embeddings. Mathematically, the conditioned latent diffusion model can be learned by optimizing the following loss: 

\begin{equation}
    L_{L D M}=\mathbb{E}_{\mathcal{E}(x), c, \epsilon, t}\left[\left\|\epsilon_\theta\left(z_t, t, c\right) - \epsilon\right\|_2^2\right]
\label{eq:ldm}
\end{equation}
where, $z_t$ is the latent version of the input $x_t$ provided by the VAE as $z = \mathcal{E}(x)$. $x_t$ is the noise added version of the input $x$, at a timestep of $t$, where $x_t = \alpha_tx_0 + (1 - \alpha_t)\epsilon$ and $\alpha_t$ decreases with the timestamp $t$. Noise is denoted by $\epsilon \sim \mathcal{N}(0,1)$. $\epsilon_\theta$ is the U-Net. Lastly, $c$ denotes the conditioning variable, and for the text guided models, it is given by processing the given text with CLIP text encoder \cite{radford2021learning}.

\subsubsection{Image Inpainting}\label{subsection:inpainting}

For the inpainting task, the objective is defined as follows: Given an image $x$, a binary map of edit region $m$ (where edit region pixels are 1), and a reference image (or images), $r$, the objective is to generate an output image, where the edited region given by $m$ is as similar as possible to $r$, and regions defined by $\mathds{1}-m$ remains as unchanged as possible, where $\mathds{1}$ denotes all ones matrix. However, the objective is not to just copy paste the given reference image in the mapped region, but to do it as plausible and realistic as possible as preserving the reference image's features is especially important in the e-commerce setting. For the inpainting, the objective can be defined mathematically by:

\begin{equation}
    L_{L D M}=\mathbb{E}_{\mathcal{E}(x), c, \epsilon, t}\left[\left\|\epsilon_\theta\left(z_t, \mathcal{E}((\mathds{1}-m) \odot x), m, t, c\right) - \epsilon\right\|_2^2\right]
\label{eq:inpaint}
\end{equation}

Here, U-Net takes two more inputs in addition to input latents, VAE processed masked image (masked latents), and the mask itself. Stable Diffusion has an inpainting model which was trained on laion-aesthetics v2.5 using classifier-free guidance \cite{ho2022classifier}, where during training, synthetic masks are generated to mask 25\% of the pixels, which in turn conditions the model for inpainting.

\subsection{Paint by Example (PBE)}\label{subsection:pbe}

The text conditioned inpainting is generally not enough to embed fine-grained details that define the reference objects especially when preserving the item's fidelity is the main priority. Thankfully, the conditioning of the latent diffusion models are not limited to textual prompts but they can also be guided by images. However, it is not straightforward to condition the diffusion models on images as the model generally tend to copy the object given in $r$ as is, instead of blending it with the $x$. More precisely, if $c$ in Eq. \ref{eq:inpaint} is selected as image, whose embeddings are given by the CLIP image encoder, the model just learns the trivial mapping function, where, $(\mathds{1}-m) \odot x + r = x$. PBE~\cite{yang2022paint} introduced a number of design choices to tackle the trivial mapping problem. Instead of utilizing all image tokens that CLIP image encoder outputs, they leveraged only the CLS token, which helps preserving semantics while preventing the trivial solution. Furthermore, they added fully connected layers to decode the CLS token, then inject it into the U-Net.





\subsection{Dreambooth}\label{subsection:dreambooth}

Instead of providing a reference image during inference time, Dreambooth~\cite{ruiz2022dreambooth} aims to inject a novel concept into the diffusion model in a few shot fine-tuning setting. This is achieved by fine-tuning the U-Net with a few reference images of the object, and a prompt in a format of ``$a [unique\, token] [class\,noun]"$, where $[unique\,token]$ is a word that does not have a strong prior in both the text encoder (e.g. a random word like $``$nbsn$"$) and the diffusion model. $[class\,noun]$ is the class of the reference images, which is used to limit the model's prior of the reference image's class. This way, the diffusion model learns this unique object and its identifier, and thus could leverage its visual prior to generate the object in novel poses on different backgrounds. This is achieved by fine-tuning Eq. \ref{eq:ldm} with a few reference images using the same conditioning vector of $``a [unique\, token] [class\,noun]"$. If the reference images are provided from different poses, it greatly affects the models ability to generate the concept in novel views. 

There are challenges with fine-tuning the entire U-Net of Stable Diffusion with a few images. In \cite{ruiz2022dreambooth}, authors identified that they had two main issues: Language-Drift and overfitting. Language-Drift is the phenomenon of associating the reference images with the given class noun. For example, if a picture of a tshirt is used as a reference with a prompt "a nbsn tshirt", then the model forgets its generalized understanding of a tshirt and associates the reference image tshirt. However, this is not really an issue in the e-commerce setting since our aim is not to preserve the models generalization capacity over the reference class, but to teach it the reference by keeping its fidelity as high as possible. The authors proposed a loss function called "class specific prior preservation loss" to help prevent overfitting. This loss function uses both the provided reference images and the model's own generated samples for a specific class noun. The purpose of this loss function is to prevent the model from forgetting how to generalize for the specific class noun, which is a problem known as "catastrophic forgetting." However, since our objective is not to keep our class token generalizable, it does not help in our case. Moreover, for the e-commerce setting it also often leads to sub-optimal results as most of the e-commerce items are of novel concepts. For example, when prompted with "roman armor" the model retrieves a lot of irrelevant images, thus using them on fine-tuning misleads to representation of the reference image.

\subsection{DreamPaint}\label{subsection:dreampaint}

It is highly likely that the textual conditioning alone is not enough to embed the high-fidelity content of the product images as product titles are not meant to fully describe the item in detail. Especially high-fidelity items can hardly be described by textual prompts only, thus it is clear that a visual reference is needed. Furthermore, pre-trained models do not have strong priors over many of the e-commerce items, as they are not represented in the bulk datasets compared to other natural images like animals, faces etc.

Using only global embeddings in PBE results in model that omit high-fidelity details of the reference image, which makes it unsuitable for the e-commerce inpainting setting, especially for the items that the model has a low prior. As buyers would like to see the item as similar as possible as given in the catalog, Dreambooth approach seems more plausible. However, the original Dreambooth does not support inpainting.

We propose to merge two pipelines for the e-commerce inpainting case. First we implemented a Masked Dreambooth model to introduce the new items to the diffusion model by providing a number of various poses of the object alongside with a new identifying token. During training, with equal probability, we mask our image latents either with rectangular or elliptic masks (since these are the most common mask shapes used by users). In addition, we generate object-shaped masks by utilizing the ClipSeg \cite{luddecke2022image} model along with the class noun of the object, the imperfections from ClipSeg segmentation mask makes our model more robust to arbitrary shaped masks. We then optimize Eq. \ref{eq:inpaint} in the dreambooth fine-tuning, and save the U-Net and Text Encoder weights. On inference time, we load the saved modules into the Stable Diffusion Inpaint Model. The high-level overview of our method is shown in Fig. \ref{fig:method}.



\begin{figure*}[t]
    \centering
    \includegraphics[width=\textwidth]{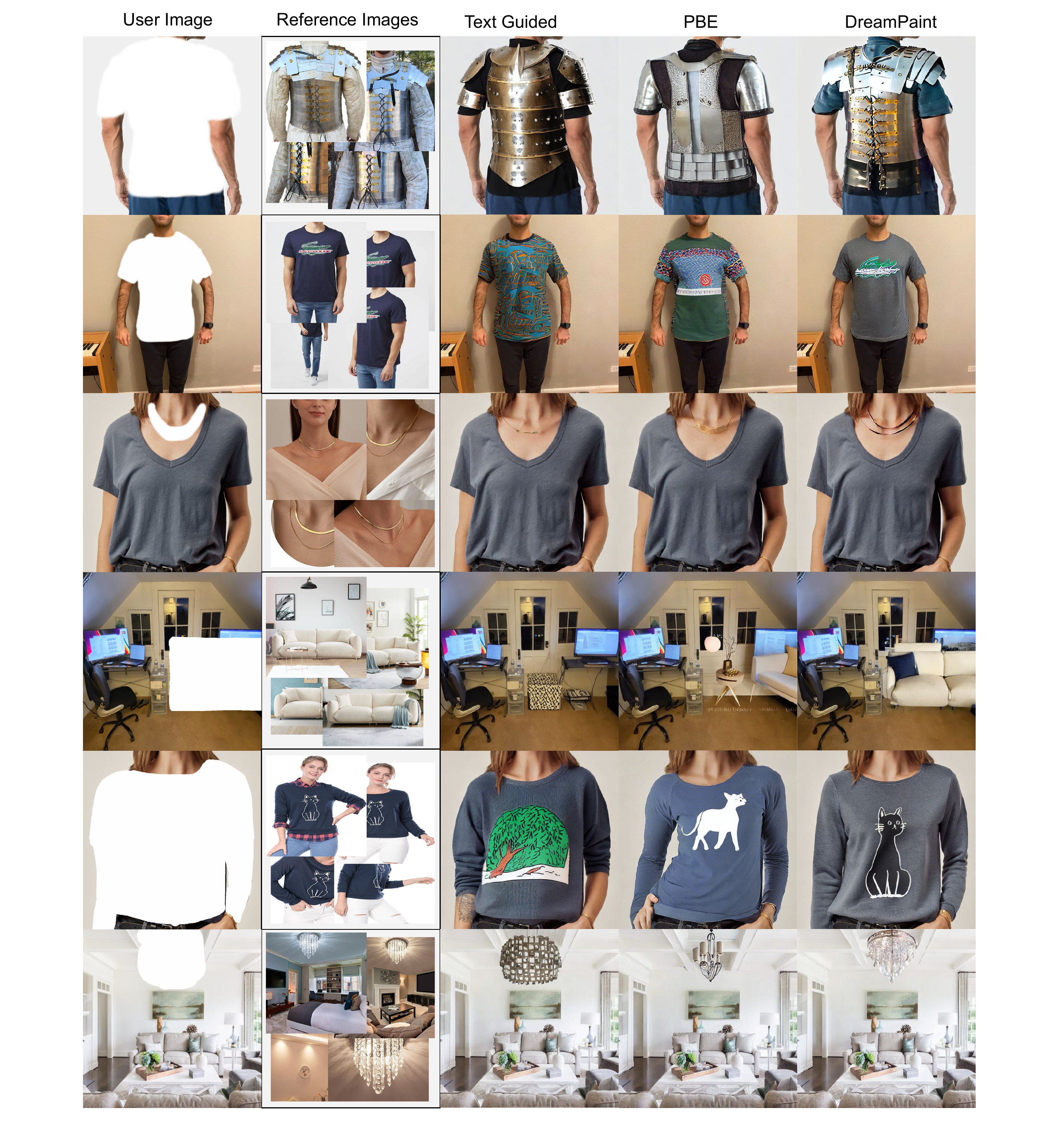}
    \caption{Comparisons against text guided and image guided diffusion models. Note that PBE and Text guided diffusion models often fail to encode the item-specific details while DreamPaint preserves most of the item specific features. }
    \label{fig:examples}
\end{figure*}
\section{Experiments}

\subsection{Implementation Details}
We utilize Stable Diffusion v1-4 as our main generative backbone. We fine-tune it for 500 steps for each item to generate 512x512 images with a learning rate of 5e-6 which is kept constant for DreamPaint. During inference, we set classifier-free guidance to 10. 
\section{Results}


\subsection{Quantitative Results}

Quantitative evaluation of the models is very challenging because our objective is to keep the object fidelity, and quantitative metrics mostly are inadequate in that regard. For example, consider two t-shirts with the same color and style, however with different logos, which only spans a couple of pixels. Fréchet Inception Distance (FID) between these two would be extremely low, since FID uses the penultimate outputs of a large neural network, which tends to generalize the given image by omitting most fine-grained details. That is why we only compute the CLIP score between the masked region and the reference images as given in \cite{yang2022paint}. For Dreampaint and Text Guided Stable Diffusion, we compared the inpainted region against all the reference images and calculated the cosine similarity between the CLIP embeddings of all (generated, reference) pairs then reported the maximum score. For PBE, we run the model for all reference images individually. We then calculated the CLIP score between (generated, reference) pairs and reported the maximum score. The results are given in Table \ref{tab:quant}. Even though PBE uses CLIP CLS token as its reference image embedder, CLIP based evaluation still favors DreamPaint.



\begin{table}
  \caption{Quantitative comparison of CLIP score between our approach and the baselines.}
  \label{tab:quant}
  \begin{tabular}{ccl}
    \toprule
    Method&CLIP Score($\uparrow)$\\
    \midrule
    SD Inpaint with Text Guidance &  $0.62$ \\
    Paint by Example &  $0.68$\\
    Ours &   \textbf{$0.70$} \\
  \bottomrule
\end{tabular}
\centering
\end{table}


\subsection{User Study}

We evaluated each model's performance by conducting a subjective user study. Participants are asked to evaluate each generated image in two different criteria, namely, generated object's similarity to the reference images, and how harmonious the generated object is without knowing which model generated the evaluated image. 10 participants are provided with 60 images, and asked to provide a score from 1 to 5 for each criteria where 1 is the best and 5 is the worst. Average scores can be seen on Table \ref{tab:subjective}.


\subsection{Qualitative Results}

Fig. \ref{fig:examples} shows some examples for the qualitative comparison of methods. As can be seen, text guided model yields the worst outputs as the representative capacity of item titles in e-commerce catalog is limited.  PBE gets some guidance from the reference image and mostly get the general theme right, but it misses the fine-grained details. On the cat sweater example, it understood  that there needs to be a cat in the middle but loses the characteristics of the design. DreamPaint's outputs, on the other hand, are the most resembling of the reference images. In some cases, it fails to get some basic features like color right (as seen on the second and fifth examples). This happens because of the context-appearance entanglement issue as mentioned in \cite{ruiz2022dreambooth}. In these examples, the reference images are all given in white background, which results in the model to build a strong prior for it and when that context is not found in the user provided masked image, the model alters some characteristics of the item mostly in the color space. 






\begin{figure*}[t]
    \centering
    \includegraphics[width=\textwidth]{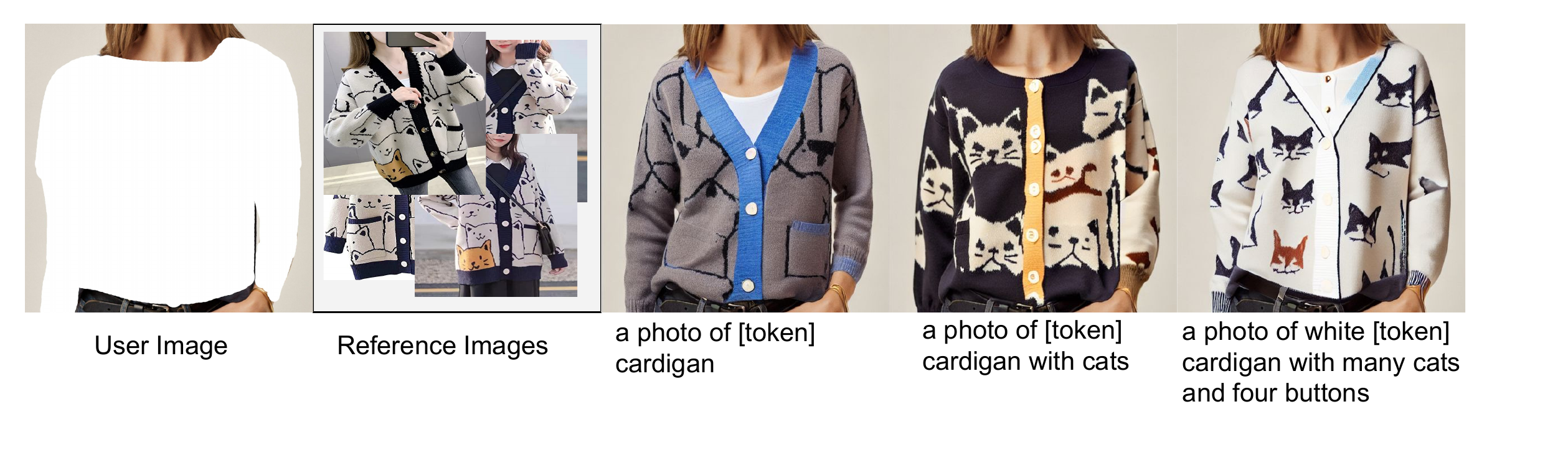}
    \caption{DreamPaint outputs with different text prompts. When trained with a photo of [token] cardigan prompt, the model misses the details of the original item. However, with additional keywords, a user can get more resembling results. }
    \label{fig:further}
\end{figure*}
\section{Ablations}

There are a number of design choices in the implementation of DreamPaint. How much does the prior preservation loss or fine-tuning the text encoder help? Can further textual guidance benefit the model in failed cases? Also how well the the model does given an unfitting mask? To answer these, we run some ablations.

\begin{table}
  \caption{Average user ratings for each method measuring similarity to the reference image and how harmoniously the generated image.}
  \label{tab:subjective}
  \begin{tabular}{ccl}
    \toprule
    Method& Similarity($\downarrow)$ &Harmony($\downarrow)$\\
    \midrule
    SD Inpaint with Text Guidance &  $4.41$ &  $2.75$ \\
    Paint by Example &  $3.82$ &  $2.57$\\
    Ours &  $2.68$ &   \textbf{$2.33$} \\
  \bottomrule
\end{tabular}
\centering
\end{table}

\textbf{Ablation Study on using Prior Preservation Loss} We found that using prior preservation loss adversely affects the model performance in the e-commerce inpainting setting, especially for rare classes that the model has little to no prior knowledge about. Thus, the model cannot retrieve meaningful class images with the class noun, and this generally results in sub-optimal injection of the new concept. Also, it often helps model to be more generalizing, but we want to preserve the high fidelity details as much as possible.

\begin{figure*}[h!]
    \centering
    \includegraphics[width=\textwidth]{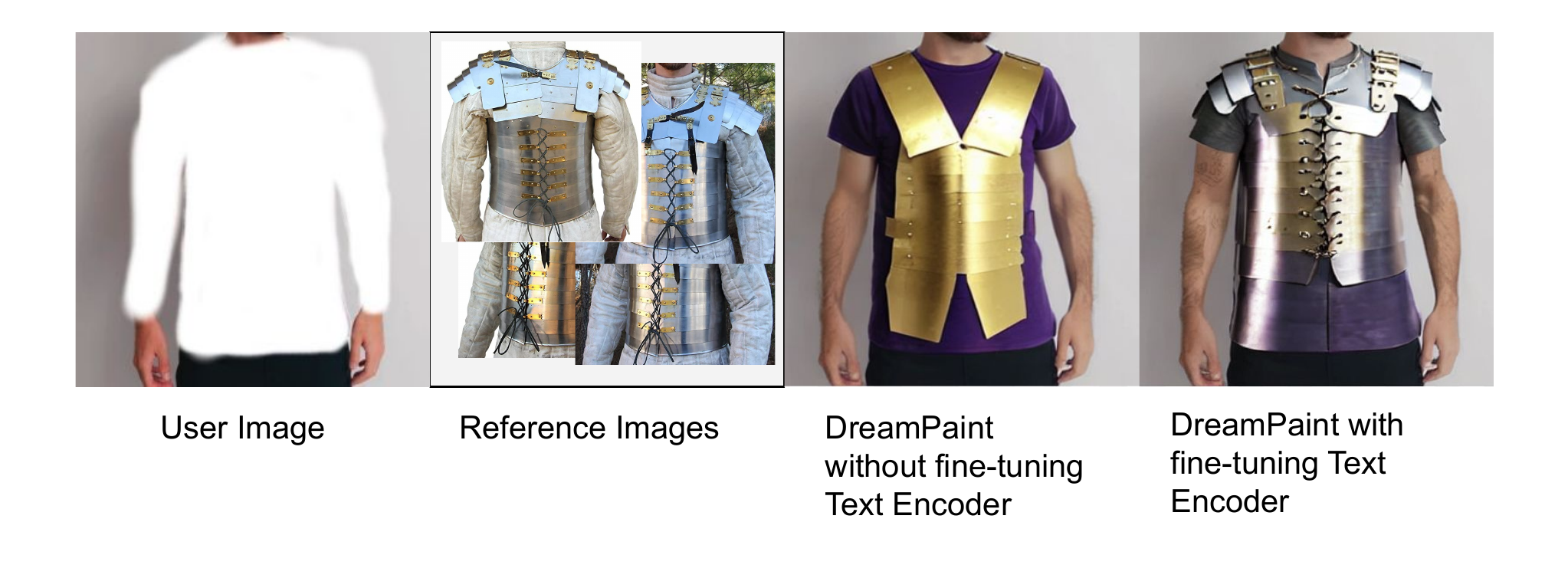}
    \caption{DreamPaint outputs with and without fine-tuning the text encoder along with the U-Net. Here we used the prompt "a [token] roman armor" }
    \label{fig:text}
\end{figure*}

\textbf{Ablation Study on Fine-tuning the Text Encoder} Stable Diffusion has a strong semantic prior for common objects, but when it comes to underrepresented tokens, such as "roman armor", it often fails in generating a specific output but uses a more generic image of an armor. We found that, if the text encoder is fine-tuned along with U-Net, it leads to optimal performance, especially when the model does not have a strong prior for that specific class noun. See Fig. \ref{fig:text} for reference.

\textbf{Ablation Study on Further Textual Guidance} One benefit of DreamPaint compared to PBE is that we are not limiting with just image guidance. After fine-tuning the model, if the results are not appealing with the generic fine-tuned prompt of "a [token] [class noun]", our model allows us to leverage text guidance to further modify the output, where providing some more context with text prompts may "reveal" the fine grained details of the newly injected concept. An example can be seen on Fig. \ref{fig:further}.

\begin{figure*}[t!]
    \centering
    \includegraphics[width=\textwidth]{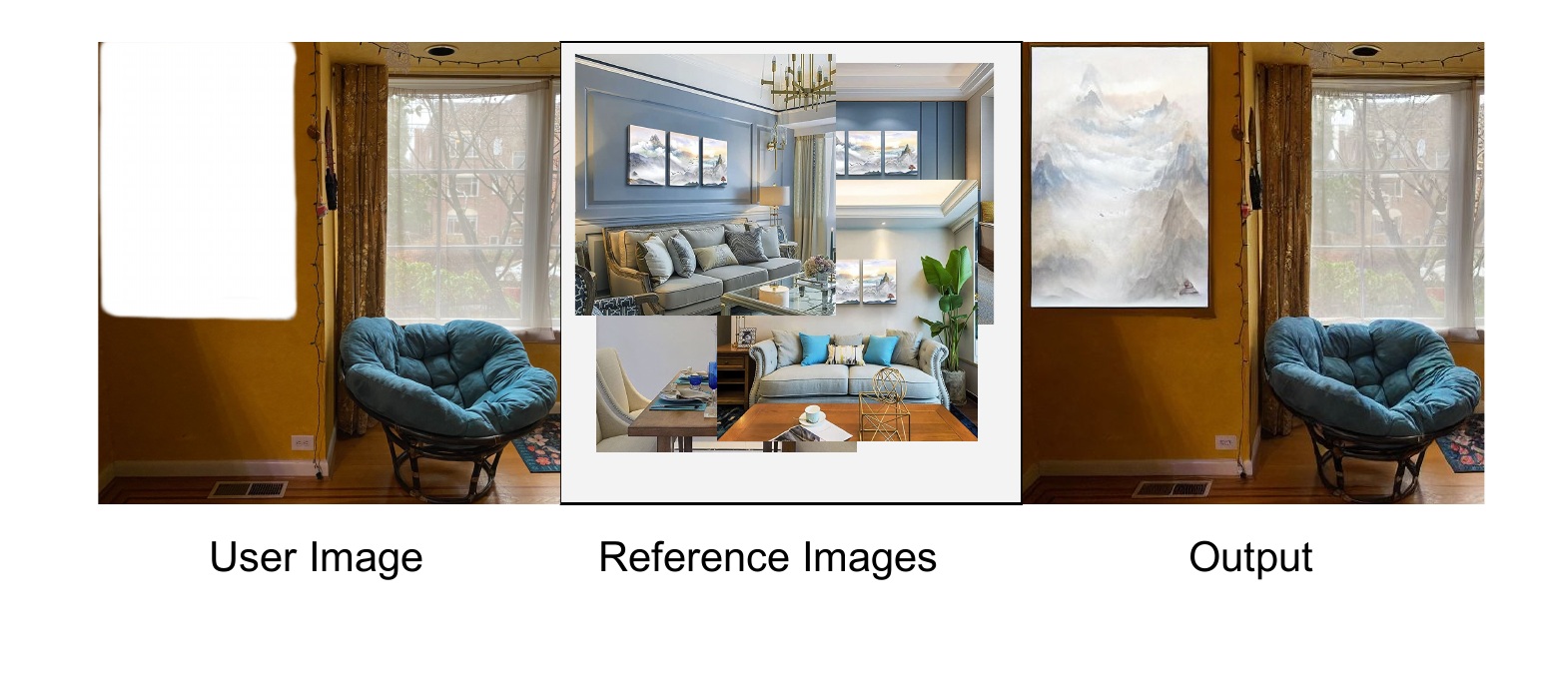}
    \caption{When given an unusually large mask size for the given object, our model model focuses on fitting it in the masked region while omitting the characteristic details of the object.}
    \label{fig:mask}
\end{figure*}

\textbf{Ablation Study on Mask Size} Our evaluation of the model's performance included subjective experiments to assess its ability to handle ill-fitting masks, which occur when the size of the given mask significantly differs from that of the object in the reference image. Results show that, in some cases, the model forcibly places the object in the masked region, while somewhat omitting its characteristic details as seen on Fig. \ref{fig:mask}, where the wall art has three pieces but here the model just created one bulky mountain art. 

\section{Conclusions}

Text guidance alone is not enough to represent e-commerce items as titles are not designed to be explicitly expressive. Therefore, it is not surprising that text guided SD inpainting fails in all metrics.

PBE works well on common objects where the model already has a strong prior, which can be exploited by a single reference image guidance. However, to prevent trivial mapping, they omit most of the high frequency signal by encoding the reference image just with the CLS token of CLIP, which results in losing fidelity in representing the reference image. Thereby, on high fidelity e-commerce items on which the model has a low prior, it yields sub-optimal results.

DreamPaint preserves highest amount of fidelity compared to text guidance and image guidance models as can be seen on both quantitative and human study metrics. It also learns a strong prior of the new item by seeing it from different angles as multiple views of an item are readily available in a e-commerce catalog. Thus, the model can learn the concept better and when given a challenging pose, it may generate a novel pose when required. Also, DreamPaint is not just image guided, but the injected token could be further detailed with extra text prompts when required. 

\subsection{Limitations}
The limitations given in the \cite{ruiz2022dreambooth} applies to DreamPaint mostly. Especially the context-appearance entanglement issue, where the appearance of the item is changing (mostly in its color) because of the given context. Moreover, depending on the mask size, the model generates its output while disregarding the physical size of the reference item in some cases. So mask input remains as a task for the user, where they have to physically measure wherever they want to inpaint the reference image on and compare it against the size of the e-commerce item given in the catalog to have realistic results. Lastly, since DreamPaint requires few-shot fine-tuning for every item, it is hard to scale it for the entire e-commerce catalog.

\bibliography{tmlr}
\bibliographystyle{tmlr}

\clearpage
\appendix

\end{document}